\documentclass[runningheads]{llncs}

\usepackage{eccv}

\usepackage{bm}
\usepackage{eccvabbrv}
\usepackage{multirow}
\usepackage{booktabs}
\usepackage{pifont}
\usepackage{utfsym}
\usepackage{graphicx}
\usepackage{booktabs}

\usepackage[accsupp]{axessibility}  

\usepackage{hyperref}

\usepackage{orcidlink}
\makeatletter
\def\thanks#1{\protected@xdef\@thanks{\@thanks
        \protect\footnotetext{#1}}}
\makeatother

\begin{document}

%
\title{Text2LiDAR: Text-guided LiDAR Point Cloud Generation via Equirectangular Transformer}

\titlerunning{Text2LiDAR}

\author{Yang Wu\inst{1} \and
Kaihua Zhang\inst{4} \and
Jianjun Qian\inst{1} \and Jin Xie\inst{2,3\dag} \and Jian Yang\inst{1}
\thanks{$\dag$ Corresponding author.}} 

\authorrunning{Y.~Wu et al.}

\institute{PCA Lab, Nanjing University of Science and Technology, Nanjing, China \and
State Key Laboratory for Novel Software Technology, Nanjing University, Nanjing, China \and
School of Intelligence Science and Technology, Nanjing University, Suzhou, China \and
B-DAT and CICAEET, Nanjing University of Information Science and Technology, Nanjing, China \\
\email{\{wuyang98,csjqian,csjxie,csjyang\}@njust.edu.cn zhkhua@gmail.com} }

\maketitle


\begin{abstract}
  The complex traffic environment and various weather conditions make the collection of LiDAR data expensive and challenging. 
  Achieving high-quality and controllable LiDAR data generation is urgently needed, controlling with text is a common practice, but there is little research in this field.
  %
  To this end, we propose Text2LiDAR, the first efficient, diverse, and text-controllable LiDAR data generation model. 
  %
  %
  Specifically, we design an equirectangular transformer architecture, utilizing the designed equirectangular attention to capture LiDAR features in a manner with data characteristics. 
  Then, we design a control-signal embedding injector to efficiently integrate control signals through the global-to-focused attention mechanism.
  Additionally, we devise a frequency modulator to assist the model in recovering high-frequency details, ensuring the clarity of the generated point cloud.
  To foster development in the field and optimize text-controlled generation performance, we construct nuLiDARtext which offers diverse text descriptors for 34,149 LiDAR point clouds from 850 scenes.
  Experiments on uncontrolled and text-controlled generation in various forms on KITTI-360 and nuScenes datasets demonstrate the superiority of our approach. The project can be found at \url{https://github.com/wuyang98/Text2LiDAR}
\end{abstract}

\section{Introduction}
LiDAR provides accurate 3D geometry and distance information about the surroundings, enabling robots to understand the 3D environment. 
This capability makes LiDAR one of the most favored sensors in various autonomous systems, such as autonomous driving~\cite{janai2020computer, cui2024survey, gulino2024waymax}, unmanned surveying~\cite{deliry2021accuracy, mohsan2023unmanned}, indoor exploration~\cite{zou2021comparative, yan2022multi, yin2023semantic}, to name a few.
However, obtaining LiDAR data is not that straightforward.
First, the price of LiDAR and its associated equipment is quite high~\cite{bakhshi2020maximizing}.
%
Second, data collection in challenging situations presents safety and ethical concerns~\cite{kong2023robo3d, piroli2023energy, dreissig2023survey}.
%
\begin{figure}[!t]
\centering
\includegraphics[width=0.9\textwidth]{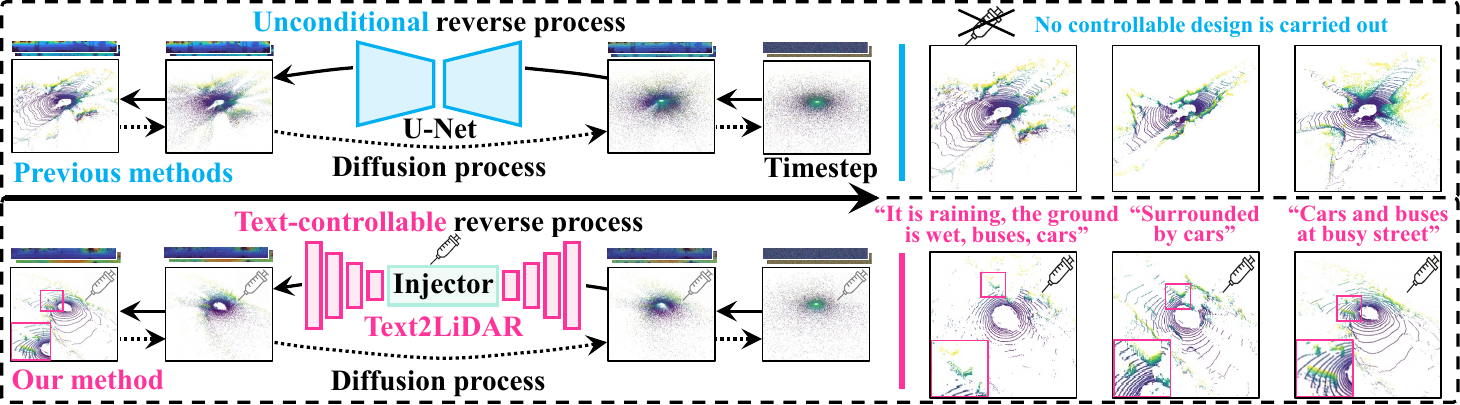}
\caption{Schematic comparison of our Text2LiDAR and the existing diffusion-based generation framework~\cite{zyrianov2022learning, nakashima2023lidar} without text guidance.}
\label{fig: introduction}
\end{figure}
Therefore, high-quality generation of LiDAR point cloud~\cite{dosovitskiy2017carla, manivasagam2020lidarsim, caccia2019deep, nakashima2021learning, nakashima2023generative, zyrianov2022learning, nakashima2023lidar} is becoming a frontier research area.

Existing works make a great effort in uncontrolled LiDAR point cloud generation that only uses single-modal LiDAR data.
CARLA~\cite{dosovitskiy2017carla} starts from the physical meaning of LiDAR and simulates the imaging process. Due to significant disparities between physical models, CARLA cannot achieve satisfying performance.
Afterward, Lidarsim~\cite{manivasagam2020lidarsim} combines a physics-based and a learning-based simulation that can achieve better generations. However, it requires scanning real scenes in advance, which is time-consuming and labor-intensive.
To address this issue, some asset-free methods~\cite{caccia2019deep, nakashima2021learning, nakashima2023generative} utilize the pure learning-based approaches, but they cannot fit more nonlinear distributions, limiting the diversity of the generation.  
Subsequent methods~\cite{zyrianov2022learning, nakashima2023lidar} achieve improved results by employing U-Net~\cite{ronneberger2015u, saharia2022photorealistic} as the denoising network in diffusion-based approaches.
As illustrated in the upper part of Figure~\ref{fig: introduction}, the diffusion process can stimulate more complicated data distributions, thereby achieving more satisfying generations.

With the introduction of the CLIP~\cite{radford2021learning} and the diffusion model~\cite{song2019generative, ho2020denoising}, the text-controlled generation tasks such as Text2image~\cite{xu2024imagereward, kim2023dense, wang2022clip} and Text2video~\cite{wang2023modelscope, wu2023tune} are developing rapidly. 
However, no works explore the text-guided paradigm in the field of LiDAR data generation.
This primarily faces two challenges: 
\textbf{(1) No controllable generation architecture specially designed for equirectangular images and text.} 
Current leading methods~\cite{zyrianov2022learning, nakashima2023lidar} all employ convolutional denoising architecture, represented by U-Net~\cite{ronneberger2015u, saharia2022photorealistic}. The convolution architecture has two main limitations:
first, ill-suiting for equirectangular images which have a circular structure, and convolutions disrupt the continuous relationship between pixels. 
Second, poor scalability, makes it inefficient and inconvenient to adapt to control signals from different modalities~\cite{hu2021lora}.
Besides, the existing methods~\cite{nakashima2023generative, zyrianov2022learning, nakashima2023lidar} also overlook the correspondence between high-frequency information in equirectangular images and the structure of point cloud objects.
These motivate us to explore how to build a unified controllable generation architecture that is compatible with the multi-modal signals of equirectangular images and text.
\textbf{(2) No reliable text-LiDAR pairs for contrastive learning.}
High-quality paired text-LiDAR data not only needs to describe the main objects in the LiDAR point cloud but also needs to include numerous diverse scenarios about weather, lighting, vehicle poses, and environmental structure to form a comprehensive description. Unfortunately, current mainstream datasets~\cite{geiger2012we, liao2022kitti, nuScenes, behley2019iccv} cannot provide high-quality paired data. How to reasonably construct high-quality text-LiDAR data pair to adapt to the rapidly evolving field is also an important issue we aim to address.

In this paper, to address the challenge \textbf{(1)}, 
we present Text2LiDAR, a high-quality, text-controllable equirectangular transformer for LiDAR point cloud generation. 
As a usual practice~\cite{nakashima2023generative, zyrianov2022learning, nakashima2023lidar}, we convert each LiDAR scan into an equirectangular image and design targeted strategies based on its characteristics.
We first design the equirectangular attention (EA) and reverse-EA (REA) for feature extraction and upsampling. They can capture long-range relationships between arbitrary two points, adapting to the circular structure of the equirectangular image, and addressing the disruption caused by convolution. Specifically, the EA introduces Fourier features to preserve 3D positional information while increasing differences between adjacent tokens for better model learning. 
Besides, the EA implements different-scale mutually overlapping unfolding operations to extract both global and local features, addressing the drastic deformations of objects caused by the elongated nature of equirectangular images.
Then, to efficiently perform control signal fusion, we design a control-signal embedding injector (CEI) through a global-to-focused attention mechanism, endowing the model with text-controllable capabilities.
At last, we design a frequency modulator (FM) to address the smooth characteristics of equirectangular images and overcome the smoothing tendency of MLP structures, ensuring the details of the generation. 
To overcome the challenge \textbf{(2)}, we construct a total of 34,149 pairs of high-quality text-LiDAR data across 850 scenes from the nuScenes~\cite{nuScenes}, dubbed as nuLiDARtext.
Based on the textual descriptions in nuScenes, we correct numerous abbreviations, spelling errors, and logical mistakes, and specifically adapt the text for LiDAR data. nuLiDARtext enhances the reliability of the text generation results and contributes to advancements in the field.
%
%
The main technical contributions are summarized as follows:
\begin{itemize}
\item We propose the first effective text-controllable LiDAR point cloud generation framework, Text2LiDAR, which fully considers and adapts to the physical characteristics of the equirectangular image.
\item We propose two novel module designs including a CEI and an FM. The CEI can progressively and robustly integrate control signals with dominant features through the global-to-focused attention mechanism, while the FM addresses the smoothing characteristics of equirectangular images and assists in model training, enhancing the quality of generation.
\item To advance the field of LiDAR point cloud generation, nuLiDARtext is constructed, comprising 34,149 pairs of text-LiDAR data across 850 scenes. 
\end{itemize}

\section{Related Work}
\noindent\textbf{Point Cloud Generation.}
LiDAR point cloud generation is a subset of point cloud generation, emphasizing the generation of point cloud in autonomous driving scenarios~\cite{kong2023robo3d, wang2023multi}. There is a strong correlation between the two tasks, due to not having to consider the surrounding scenes, the point cloud generation has earlier and more extensive related research.
Earlier methods~\cite{dosovitskiy2017carla, manivasagam2020lidarsim} often rely on physical models, which makes them constrained by LiDAR equipment and capable only of achieving coarse generation. 
Sever works~\cite{achlioptas2018learning, valsesia2018learning, zamorski2020adversarial, caccia2019deep, klokov2020discrete, hui2020progressive} utilize the representative generative models such as generative adversarial networks (GANs)~\cite{good2014gener} and variational autoencoders (VAEs)~\cite{kingma2013auto} to solve point cloud generation.
In addition, diverse generation methods have been proposed, achieving some effectiveness.
Wu $et \ al.$~\cite{wen2021learning} design a dual-generator framework that progressively extends the traditional GAN. 
SnowflakeNet~\cite{xiang2022snowflake} models the generation process as a snowflake-like growth, each point is generated from the original point after snowflake point deconvolution.
Pointflow~\cite{yang2019pointflow} introduces a two-level distribution structure, different levels represent different types of knowledge, allowing the model to sample point clouds of different sizes.
Lou $et \ al.$~\cite{luo2021diffusion} utilize diffusion techniques for generating point clouds, enabling high-quality point clouds with diverse scales.
Gecco~\cite{tyszkiewicz2023gecco} improves the geometric consistency of point clouds by projecting sparse image features into the point cloud and using them as conditioning during the denoising process.
By utilizing distillation techniques~\cite{hinton2015distilling, li2023curriculum, li2024promptkd}, Wu $et \ al.$~\cite{wu2023fast} shorten the straight path to a single step, shorting the generation time of the standard diffusion model.

The outdoor LiDAR point cloud is very irregular and sparse.
Due to the correlation between depth and LiDAR~\cite{milioto2019rangenet++, yan2022rignet, yan2024tri}, in the tasks related to autonomous driving, LiDAR point cloud is often converted into equirectangular images to overcome the unstructuredness and sparsity of LiDAR point cloud~\cite{meyer2019lasernet, chai2021point, milioto2019rangenet++}.
Nakashima $et \ al.$~\cite{nakashima2021learning, nakashima2023generative} decompose the noised equirectangular images into denoised forms and their corresponding dropout probabilities, significantly improving the performance.
LiDARGen~\cite{zyrianov2022learning} designs a masking strategy to simulate the ray-drop in LiDAR, achieving diverse size generation results, and verifying the feasibility of using diffusion models for LiDAR data generation.
The current leading work, R2DM~\cite{nakashima2023lidar}, designs a more mature diffusion framework and achieves significant performance improvement.
Despite numerous advancements, the realism of generated LiDAR point clouds and their diversity remain relatively low, and the absence of an efficient architecture with strong feature fusion capabilities for text control is still an unresolved issue.

\noindent\textbf{Text in Vision.}
Thanks to the massive paired image-text data and the clever and concise model design, CLIP~\cite{radford2021learning} can provide the semantically rich joint text-image representation, demonstrates strong capabilities across numerous visual tasks, such as low-light image enhancement~\cite{xue2024low}, open-vocabulary object detection~\cite{kuo2022f}, image style transfer~\cite{xu2024spectralclip}.
Thanks to the abundance of large-scale paired text-image datasets~\cite{schuhmann2022laion, lin2014microsoft, wang2022diffusiondb, cho2023dall, gupta2019lvis} and the cross-modal representation capability of CLIP, the currently booming field of text-controlled image generation~\cite{xu2024imagereward, kim2023dense, wang2022clip, ramesh2022hierarchical, crowson2022vqgan, zhou2022towards, ge2023expressive, saharia2022photorealistic} effectively leverage text embeddings as control signals to guide the entire image generation process and achieve amazing results. 
All of these demonstrate that text and image features can be effectively fused. Thanks to the training objectives involving text-image pairs, the text embeddings generated by CLIP possess richer semantic information, making them particularly suitable as control conditions for generation tasks~\cite{saharia2022photorealistic}.

In the 3D vision domain, there is a scarcity of paired datasets, and compared to image generation, the development of 3D generation is relatively slow. 
Some methods~\cite{wu2023sketch, poole2022dreamfusion, chen2023fantasia3d, chen2023text, kasten2024point} leverage pre-trained models and utilize Nerf~\cite{mildenhall2021nerf} or Gaussian splatting~\cite{kerbl20233d} to achieve 3D generation.
Nevertheless, in complex, realistic, and diverse 3D environments, these approaches are not universally applicable to all 3D generation tasks. Constructing text-paired datasets tailored to specific 3D generation tasks and conducting research on this basis is therefore particularly urgent.
Some efforts are made in this direction. Chen $et \ al.$~\cite{chen2019text2shape} construct a text-shape dataset and achieve text-controlled shape generation based on it. Liu $et \ al.$~\cite{liu2022towards} further improve this work by decoupling the shape and color predictions for learning features.
However, there are currently no relevant text-LiDAR datasets and generation frameworks. This paper aims to address these issues.

\section{Method}
\subsection{Preliminary}
This section introduces the formulation of the denoising diffusion probabilistic model (DDPM) and the loss function.
%
%
As shown in Figure~\ref{fig: introduction}, the DDPM employs a forward diffusion process to gradually destroy the data sample $\textbf{x}$ by adding noise as evolving the timestep $t \in [0,1]$ until it becomes pure Gaussian noise. It also contains a backward reverse process, which aims at predicting the noise in each timestep and converting the pure Gaussian noise back into the data $\textbf{x}$.
To be more specific, at the timestep $t$, we can obtain the noised sample $\textbf{x}_t$ through $q(\textbf{x}_{t}|\textbf{x}) = \mathcal{N}(\alpha_{t}\textbf{x}, \sigma^{2}_{t}\textbf{I}),$
where $\textbf{x}_t$ can be re-parameterized as: $\textbf{x}_t = \alpha_{t}\textbf{x} + \sigma_{t}\bm{\epsilon}_t$,
$\bm{\epsilon}_{t} \sim \mathcal{N}(\textbf{0},\textbf{I})$ and $\bm{\epsilon}_t$ is the noise  that vary with timestep $t$. $\alpha_t$ and $\sigma_t$ are hyperparameters that depend on timestep $t$ following the $\alpha$-cosine schedule~\cite{nakashima2023lidar}, we set $\alpha_t = \texttt{cos}(\pi t/2)$, $\sigma_t = \texttt{sin}(\pi t/2)$. Under the assumption $\alpha^{2}_{t} + \sigma^{2}_{t} = 1$, the process of obtaining the intermediate noised sample $x_s$ can be described as $q(\textbf{x}_{t}|\textbf{x}_{s}) = \mathcal{N}(\alpha_{t|s}\textbf{x}_s, \sigma^{2}_{t|s}\textbf{I}),$
where $0 \leq s < t \leq 1$, $\alpha_{t|s} = \alpha_{t} / \alpha_{s}$ and $\sigma^{2}_{t|s} = \sigma_{t} - \alpha_{t|s}\sigma_{s}$. The corresponding reverse process can be described as:
\begin{equation}
p(\textbf{x}_{s}|\textbf{x}_{t}) = q(\textbf{x}_{s}|\textbf{x}_{t},\textbf{x}).
\label{eq: x_last}
\end{equation}
After obtaining the noised sample $\textbf{x}_t$, we need to design a denoiser $\texttt{Text2LiDAR}_{\bm{\varphi}}$ to predict the noise $\hat{\bm{\epsilon}}_{t} = \texttt{Text2LiDAR}_{\bm{\varphi}}(\textbf{x}_{t}, t)$ at each timestep $t$. Then, the denoised results can be obtained through Equation~\ref{eq: x_last}. Completing the entire denoising process for each timestep $t$, we can yield the final generated result. We use the mean squared error (MSE) loss function for the training process:
\begin{equation}
\mathcal{L} = \mathbb{E}_{\textbf{x},\bm{\epsilon} \sim \mathcal{N}(\textbf{0},\textbf{I}),t} [||\bm{\epsilon}_{t} - \texttt{Text2LiDAR}_{\bm{\varphi}}(\textbf{x}_{t},t)||^{2}_{2}],
\label{eq: loss}
\end{equation}
where $\bm{\varphi}$ means the learnable parameters. As is customary~\cite{nakashima2023lidar}, our denoiser is also conditioned on $t$. After training, we can obtain the final generated results by recursively evaluating $p(\textbf{x}_{s}|\textbf{x}_{t})$ through the process for $t = 1 \rightarrow 0$.

\subsection{Text2LiDAR Denoising Network}
\begin{figure}[!t]
\includegraphics[width=1\textwidth]{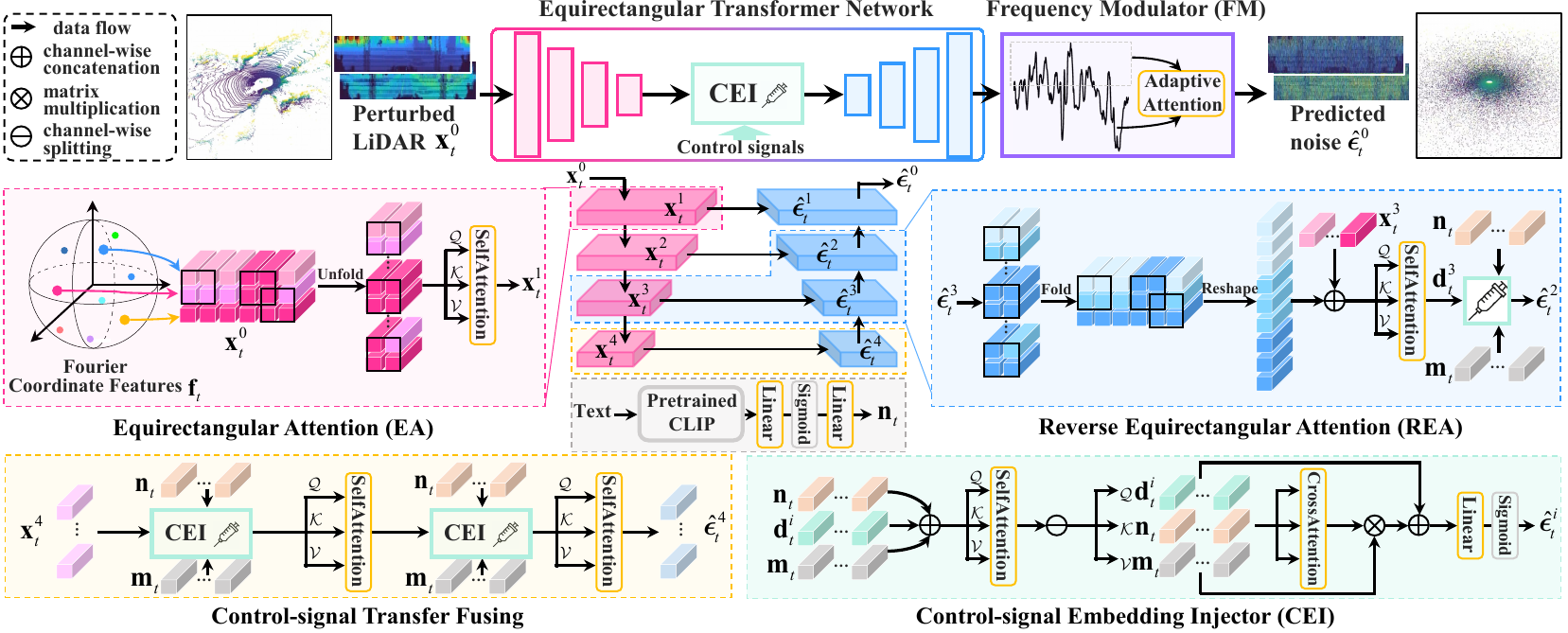}
\caption{The architecture of the designed Text2LiDAR, 
where the designed equirectangular transformer is composed of stacked EA (encoding stage) and REA (decoding stage).
The feature sequence will start interacting with the control signal at the 4th layer and be fed into a 4-layer decoder composed of REA. 
During decoding, the feature sequence continuously fuses the control signal through CEI. 
Finally, after frequency modulation, we can get the predicted noise.}
\label{fig: flow}
\end{figure}

Figure~\ref{fig: flow} illustrates the architecture of our Text2LiDAR denoising network.
At each timestep, the Text2LiDAR takes a noised equirectangular image $\textbf{x}_{t}^0\in\mathbb{R}^{H\times W\times 2}$ as input and outputs the predicted noise $\hat{\bm{\epsilon}}_{t}^0$ with the same size. 
The entire process is mainly composed of three parts, an equirectangular transformer network, a control-signal embedding injector, and a frequency modulator.

\noindent\textbf{Equirectangular Transformer Network.}
Unlike regular images, equirectangular images possess unique physical properties. To this end, we design the equirectangular attention (EA) and highly adapt equirectangular images in three aspects.
First, equirectangular images have a circular structure, previous methods employing circular convolution~\cite{schubert2019circular} still have limitations in expanding the convolution boundaries. In contrast, we utilize self-attention to break through these boundaries, enabling the capture of long-range relationships between arbitrary points.
Second, pixels in the equirectangular image correspond to 3D positions. Previous methods only treated angular coordinates as additional input, resulting in minimal differences for adjacent regions, which is unfavorable for learning with MLP structures. Therefore, we utilize Fourier features~\cite{tancik2020fourier} and extend the elevation and azimuth angles to frequency components of powers of two. This preserves the 3D priors while magnifying differences between neighboring positions, facilitating better model learning.
Third, due to the elongated nature of the equirectangular image, the targets within it can undergo particularly exaggerated scale variations based on their distances, which are overlooked by previous methods. To address this, we use the mutually overlapping unfolding to cut the input sequence into different scales at different stages for local modeling. On these bases, our model can incorporate physical meanings to conduct feature extraction comprehensively.
The process can be written as:
\begin{equation}
\textbf{x}_t^{i+1} = \texttt{EA}(\textbf{x}_t^{i}) = \texttt{SelfAttention}(\texttt{Unfold}(\textbf{x}_t^{i}\oplus \textbf{f}_t)),
\label{eq: locao-global}
\end{equation}
through the same steps, we can progressively obtain multi-level feature embeddings $\textbf{x}^{i}_{t} \in \mathbb{R}^{H/2^{i} \times W/2^{i} \times C}, i = [1,2,3,4]$.

Before decoding, as shown in the bottom-left corner of Figure~\ref{fig: flow}, we process $\textbf{x}_t^4$ using the control-signal transfer fusing for an initial fusion of control signals. 
In the decoding part of Text2LiDAR, we design reverse equirectangular attention (REA) for upsampling, allowing the continued capture of global-to-local relationships. 
For better recovering object details, we introduce features from the encoding stage. 
Simultaneously, to enhance the guidance of the embedding on the model, we use the designed control-signal embedding injector (CEI) at each upsampling stage to provide control.
Through four stages of upsampling, we can upsample the token sequences to high resolution, matching the input size. This process can be written as:
\begin{equation}
\hat{\bm{\epsilon}}_t^i=\texttt{REA}(\textbf{x}_t^{i+1},\hat{\bm{\epsilon}}_t^{i+1}, \textbf{n}_t, \textbf{m}_t)=\texttt{CEI}(\texttt{SelfAttention}(\textbf{x}_t^{i+1}\oplus\texttt{Fold}(\hat{\bm{\epsilon}}_t^{i+1})), \textbf{n}_t,\textbf{m}_t),
\label{eq: local-global}
\end{equation}
where $\textbf{n}_t$ is the text embedding, $\textbf{m}_t$ is the timestep embedding.

\begin{figure}[!t]
\includegraphics[width=1.0\textwidth]{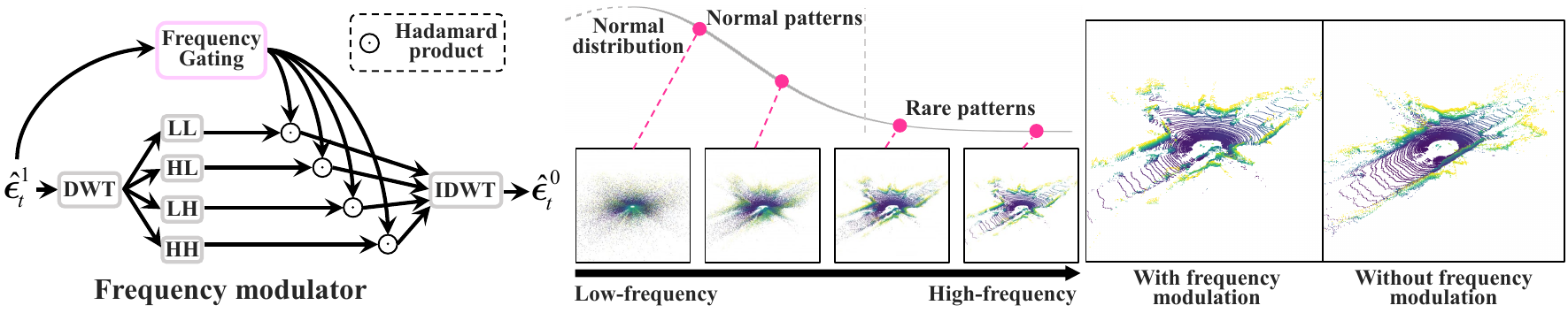}
\caption{The architecture of the frequency modulator.}
\label{fig: frequency}
\end{figure}
\noindent\textbf{Control-signal Embedding Injector.}
One major reason previous methods can't achieve text-controlled generation is the inability to unify multi-modal features into tokens and integrate them.
The bottom-right of Figure~\ref{fig: flow} illustrates the details of the CEI.
Specifically, the dominant feature sequence $\textbf{d}^{i}_{t}$ obtained from the preceding steps, text embedding $\textbf{n}_{t}$ and timestep embedding $\textbf{m}_{t}$ are first concatenated for self-attention operation, and the obtained feature sequence is then split in the order of concatenation to form sever new feature sequences $\mathcal{Q}\textbf{d}^{i}_{t}, \mathcal{K}\textbf{n}_{t}, \mathcal{V}\textbf{m}_{t}$. 
These new sequences are for cross-attention operation~\cite{vaswani2017attention, yuan2021tokens}. 
To preserve the guiding effect of the timestep $\mathcal{V}\textbf{m}_{t}$, we performed additional matrix multiplication for it.
To consolidate the dominant denoising feature sequence $\textbf{d}^{i}_{t}$, we add it to the feature sequence after multiplication, obtaining the final sequence $\hat{\bm{\epsilon}}^{i}_{t}$. This process can be formulated as:
\begin{equation}
\hat{\bm{\epsilon}}_t^i = \texttt{CrossAttention}(\texttt{Split}(\texttt{SelfAttention}(\textbf{n}_t\oplus \textbf{d}_t^i\oplus \textbf{m}_t)))\otimes\mathcal{V}\textbf{m}_t\oplus\mathcal{Q}\textbf{d}_t^i,
\label{eq: global-to-focused}
\end{equation}
This is a global-to-focused mechanism that, while integrating control signals, maintains the ability to generate key features from LiDAR data. It's worth mentioning that our designed injector can easily remove text embeddings. Only the composition of $\mathcal{QKV}$ needs to be adjusted.

\noindent\textbf{Frequency Modulator.}
As shown in Figure~\ref{fig: frequency}, in the diffusion model, low-frequency information is restored first, followed by the gradual restoration of high-frequency information~\cite{yang2023diffusion}. 
The high-frequency information in equirectangular images affects generation details, but it is prone to being smoothed out by MLP operations
To this end, we design a frequency modulator (FM) to allow the model to adaptively focus on high-frequency information. 
The left side of Figure~\ref{fig: frequency} illustrates its structure, mainly composed of a discrete wavelet transform (DWT), a frequency gating function (FG) composed of convolutions, and an inverse discrete wavelet transform (IDWT)~\cite{yang2023diffusion, fu2021dw}. 
The goal is to decompose the input $\hat{\bm{\epsilon}}_t^1$ into multi-angle high-frequency wavelet bands for modulation, guiding the model to adapt adaptively to different frequencies, alleviating the transition smoothness of the equirectangular image.
The process can be written as:
\begin{equation}
\hat{\bm{\epsilon}}_t^0 = \texttt{FM}(\hat{\bm{\epsilon}}_t^1) = \texttt{IDWT}(\texttt{DWT}(\hat{\bm{\epsilon}}_t^1)\odot \texttt{FG}(\hat{\bm{\epsilon}}_t^1)).
\label{eq: frequency}
\end{equation}
We can see from the right side of Figure~\ref{fig: frequency}, that with the designed FM, the details of the LiDAR point clouds are clearer.
In the end, we obtain the predicted denoised results and perform end-to-end model training using Equation~\ref{eq: loss}.

\begin{figure}[!t]
\centering
\includegraphics[width=1.0\textwidth]{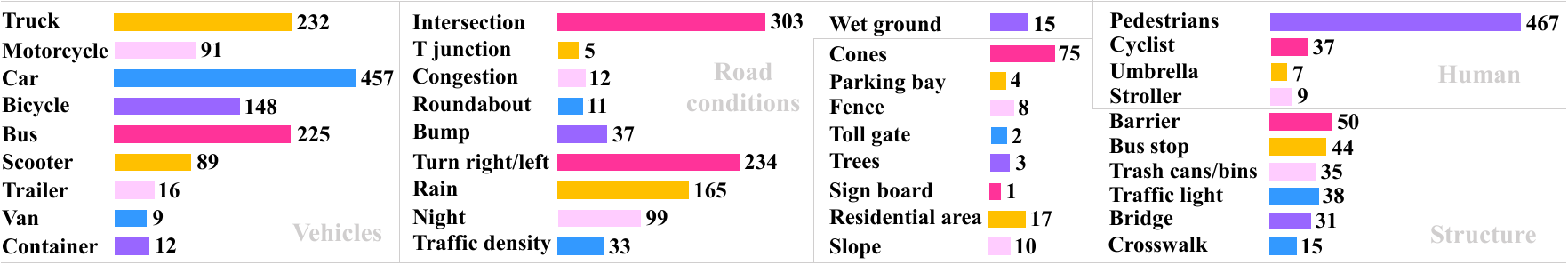}
\caption{The number of occurrences of text in 850 scenes.}
\label{fig: textdata}
\end{figure}
\subsection{Construction of NuLiDARtext}
Enabling text control can significantly enhance the practicality of LiDAR point clouds generation, meeting personalized customization needs for weather and road conditions. 
We introduce nuLiDARtext to promote the advancement of the field.
To save resources and costs, we construct text descriptions suitable for single-frame LiDAR point cloud generation on existing nuScenes~\cite{nuScenes}. 
%
The text descriptions in the nuScenes dataset are intended for describing scenes within a short period and are not paired specifically for LiDAR data.
When attempting to pair text data with LiDAR data, we find issues such as misspellings, semantic redundancies, continuous state descriptions, and interference words. Therefore, we manually adjust the descriptions of 34,149 frames across 850 scenes provided by nuScenes, including operations such as addition, deletion, modification, and standardization.
For example, we correct "ped" to "pedestrians" since this abbreviation is not conducive to obtaining effective text embedding. We also adjust instances where both "turn left" and "turn right" are simultaneously mentioned to more realistically represent "turn left". Additionally, we remove "hidden ped", which represents information not perceivable by LiDAR. We modify "waiting at the intersection" to "at the intersection", as "waiting" is a continuous state description that could dilute the effective information for a single-frame generation. We make corrections and modifications to almost every text description.
The proposed nuLiDARtext can better provide paired text-LiDAR data, promoting advancements in the field. Figure~\ref{fig: textdata} illustrates the main components of the current textual prompts, and the presentation of the dataset can be found in the supplementary materials.

\section{Experiments}
\subsection{Implementation Details}
Our model is built using PyTorch~\cite{paszke2019pytorch} and trained on 4 NVIDIA RTX 3090 GPUs, requiring a total of 300,000 training steps. 
%
%
Our model is optimized using the Adam algorithm~\cite{kingma2014adam} with exponential decay rates of 0.9 and 0.99, and the learning rate is set to $1e-4$. 

\subsection{Datasets and Evaluation Metrics}
We conduct experiments on two challenging datasets, KITTI-360 and nuScenes. 
KITTI-360~\cite{liao2022kitti} provides 360-degree, 64-beam LiDAR point clouds, allowing intelligent devices to comprehensively understand the 3D structure of the surroundings.
As usual practice~\cite{zyrianov2022learning, nakashima2023lidar}, we split KITTI-360 into two parts, with one portion (50348 frames) used for training and validation, and the other portion (26367 frames) reserved for testing. When projecting onto equirectangular images, we set the dimensions of the image to $64\times1024\times2$.
NuScenes~\cite{nuScenes} provides continuous data collection for 20 seconds, and we primarily utilize its "samples" data, which consists of 34,149 frames of 360-degree, 32-beam LiDAR point clouds.
The projected equirectangular images are set to $32\times1024\times2$.

Consistent with LiDARGen~\cite{zyrianov2022learning} and R2DM~\cite{nakashima2023lidar}, for unconditional generation, we compute the distributional dissimilarity between 10,000 generated samples and real samples, conducting evaluations in three data formats: image, point clouds, and bird's eye view (BEV).
In equirectangular image form, we utilize the pre-trained RangeNet~\cite{milioto2019rangenet} to calculate the Frechet range distance (FRD)~\cite{zyrianov2022learning}. 
In point cloud form, we use the pre-trained PointNet~\cite{qi2017pointnet} to obtain the Frechet point cloud distance (FPD) metric, which calculates similarly to FRD, to evaluate the difference between generated samples and real samples at the point cloud level.
In BEV form, we use Jensen–Shannon divergence (JSD) and minimum matching distance (MMD) to measure the distance between the marginal distributions of BEV occupancy grids.

\subsection{LiDAR Uncontrolled Generation}
\begin{figure}[!t]
\includegraphics[width=1\textwidth]{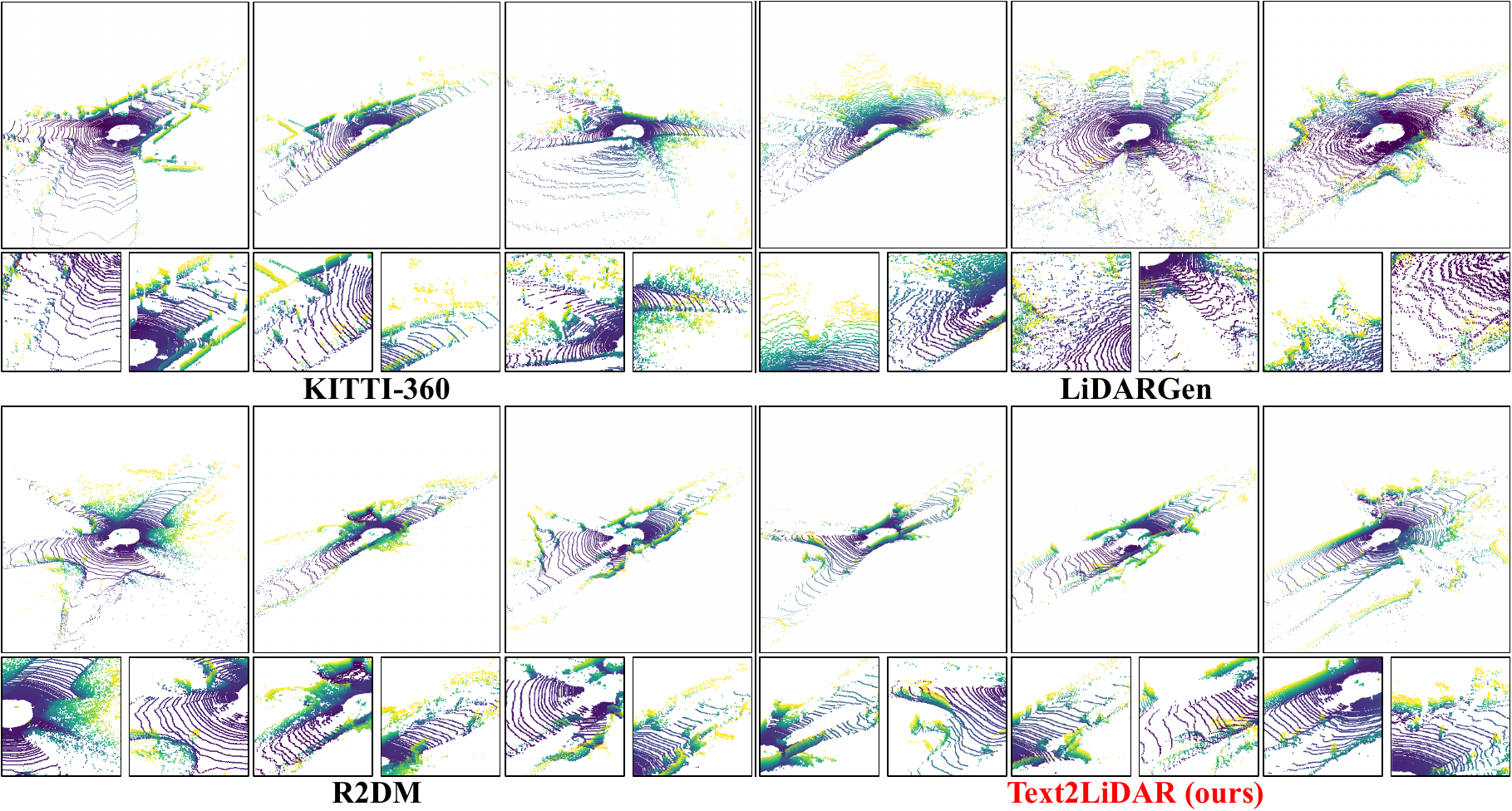}
\caption{Comparison with LiDARGen and R2DM on uncontrolled generation.}
\label{fig: unconditional}
\end{figure}
\begin{table}[!t]
\centering
\caption{Comparison of four metrics with state-of-the-art methods on KITTI-360.}
\begin{tabular}{ccccccc}
\bottomrule
\multirow{2}{*}{Method} & \multirow{2}{*}{Base} & \multirow{2}{*}{T} & Point cloud & Image  & \multicolumn{2}{c}{BEV occupancy grid} \\ \cmidrule(r){4-4}\cmidrule(r){5-5}\cmidrule{6-7} 
 & & & FPD\scriptsize$\downarrow$ & FRD\scriptsize$\downarrow$ & MMD\scriptsize$\times10^{-4}\downarrow$\qquad & JSD\scriptsize$\times10^{-2}\downarrow$ \\ \hline
LiDARGAN$^*$~\cite{caccia2019deep} & GAN & None & -  & 3003.8 & \quad 30.60 \qquad  & - \\
LiDARVAE$^*$~\cite{caccia2019deep} & VAE & None  & -  & 2261.5 & \quad 10.00 \qquad  & 16.10 \\
ProjectedGAN$^*$~\cite{sauer2021projected} & GAN & None & -  & 2117.2 & \quad 3.47 \qquad  & 8.50 \\
LiDARGen~\cite{zyrianov2022learning} & NCSNv2 & 1160  & 90.29 & 579.39 & \quad 7.39 \qquad & 7.38 \\
R2DM~\cite{nakashima2023lidar}   & DDPM   & 256   & 6.24 & \textbf{149.66} &  \quad 1.91 \qquad   & 3.05 \\
Text2LiDAR (ours)     & DDPM   & 256   & \textbf{4.81} & 164.16 & \quad \textbf{0.49} \qquad & \textbf{2.01} \\ \bottomrule
\label{tab: metrics}
\end{tabular}
\end{table}
In this section, we conducted quantitative and qualitative analyses of the results of uncontrolled generation. We compare our approach with three GAN~\cite{caccia2019deep, sauer2021projected} or VAE methods~\cite{caccia2019deep} and two diffusion methods~\cite{zyrianov2022learning, nakashima2023lidar}. 
Among them, the evaluation metrics for methods marked with "*" are provided by LiDARGen~\cite{zyrianov2022learning}, while other methods are trained with uniform settings from their source code and evaluated accordingly, and we standardize R2DM to be trained in single precision to keep the same with LiDARGen and Text2LiDAR. 

Table~\ref{tab: metrics} presents the experimental results. It can be observed that our approach achieves the best performance in terms of FPD, MMD, and JSD. This indicates that the point clouds generated by our model exhibit the most realistic distribution in both the 3D and 2D planes. 
Our method also performs well in the FRD metric, with a slight difference compared to R2DM. 
%
%
However, this does not affect our ability to generate high-quality LiDAR point clouds.

Figure~\ref{fig: unconditional} displays the uncontrolled generation results of our method and the compared methods. We can see that LiDARGen exhibits the basic characteristics of LiDAR point clouds but suffers from noticeable blurriness and more noise. 
Compared to R2DM, our method better captures the realistic state of distant beams, owing to the effective capturing of the LiDAR features from equirectangular images by our designed EA.
Thanks to the designed frequency modulator, the generation of target outlines and boundaries is also clearer in our method. 
We present more generated results in supplementary materials

\subsection{LiDAR Densification}
\begin{figure}[!t]
\centering
\includegraphics[width=0.95\textwidth]{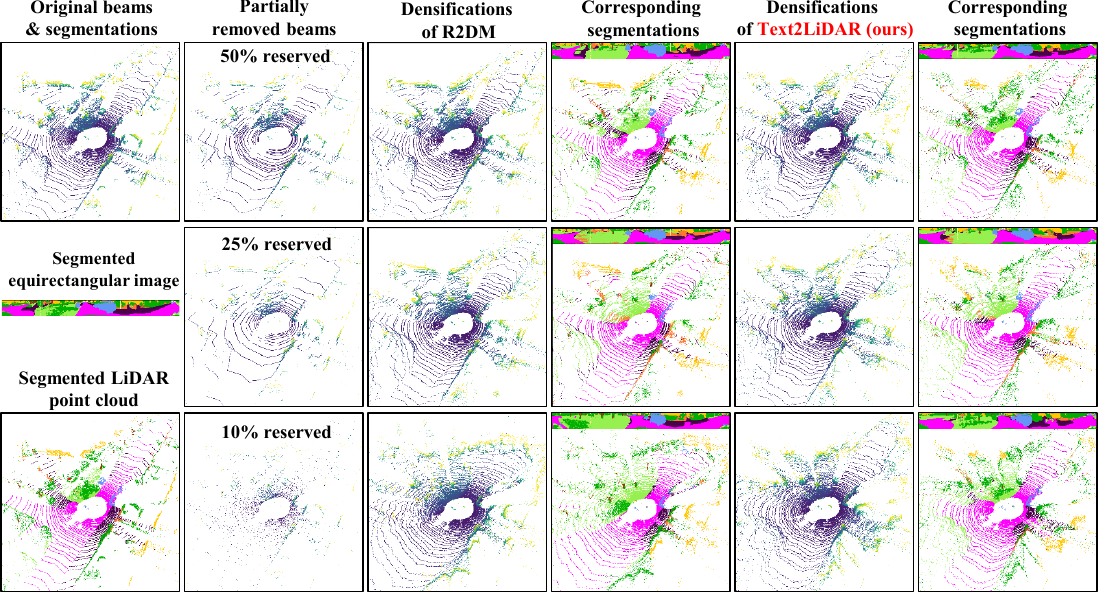}
\caption{Comparison with R2DM on densification.}
\label{fig: conditional}
\end{figure}
\begin{table}[!t]
\centering
\caption{Comparison of densification performance with R2DM on KITTI-360.}
\resizebox{0.6\textwidth}{!}{
\begin{tabular}{cccccc}
\bottomrule
\multirow{2}{*}{Method} & \multicolumn{2}{c}{Depth} & \multicolumn{2}{c}{Intensity} & Semantic \\  \cmidrule(r){2-3}\cmidrule(r){4-5}\cmidrule{6-6}
 & MAE\scriptsize$\downarrow$ & RMSE\scriptsize$\downarrow$ & MAE\scriptsize$\downarrow$ & RMSE\scriptsize$\downarrow$ & IoU\scriptsize$\% \uparrow$ \\ \hline
R2DM & 0.039 & 0.114 & 0.090 & 0.146 & 40.16 \\
Text2LiDAR(ours) & 0.035 & 0.102 & 0.073 & 0.134 & 41.37 \\ \bottomrule
\label{tab: mae}
\end{tabular}}
\end{table}
Compared to directly generating LiDAR point clouds in uncontrolled manners, densifying existing sparse LiDAR point clouds (32,16 beams) is also an effective method for data generation. It can significantly reduce data collection costs, and can effectively enhance the accuracy and safety of unmanned systems~\cite{liao2022kitti}. 
To maintain consistency with the baseline, we use the framework designed based on R2DM~\cite{nakashima2023lidar} on top of Repaint~\cite{lugmayr2022repaint} for densification.
In addition to observing the completeness and realism of the LiDAR point clouds, we also need to verify the rationality. 
For effectively evaluating, we employ RangeNet~\cite{milioto2019rangenet} for semantic segmentation on both the initial and the densification, aiming to observe whether the consistency in semantic information has been well preserved.

As shown in Figure~\ref{fig: conditional}, in the first column, we present the original LiDAR point clouds and their segmentation results. In the second column, we showcase the states of the original LiDAR point cloud with a reduction of $50\%$ and $75\%$ of beams, as well as the state with $90\%$ of points randomly removed. It is not difficult to see that the latter two are quite challenging.

In the third and fourth columns of Figure~\ref{fig: conditional}, we show the results of densification and semantic segmentation of R2DM, and in the next two columns, we present the results of our Text2LiDAR.
It is evident that the point clouds densified by our method align more closely with the characteristics of the real, and the generation of semantic information is more complete and accurate. Taking the most challenging case in the third row as an example, R2DM exhibits large-scale misjudgments for the road below and on the left, as well as misjudgments for the vegetation on the left. In contrast, our method better preserves the road surface and vegetation, aligning more closely with the characteristics of initial point clouds.

For quantitative comparison, we test $100$ samples on KITTI-360. We utilize mean absolute error (MAE) and root mean squared error (RMSE) as metrics to assess the quality of isotropic equirectangular image densification. Additionally, intersection over union (IoU) is employed to measure the ability to preserve semantic information.
Table~\ref{tab: mae} shows the performance comparison between our method and R2DM under the condition of randomly removing $90\%$ of points. We can see that our method outperforms R2DM across all metrics. This is attributed to the designed EA and REA, which perform feature extraction and upsampling based on the physical characteristics of the equirectangular image

\subsection{Text-controlled Generation}
\begin{figure}[!t]
\centering
\includegraphics[width=0.95\textwidth]{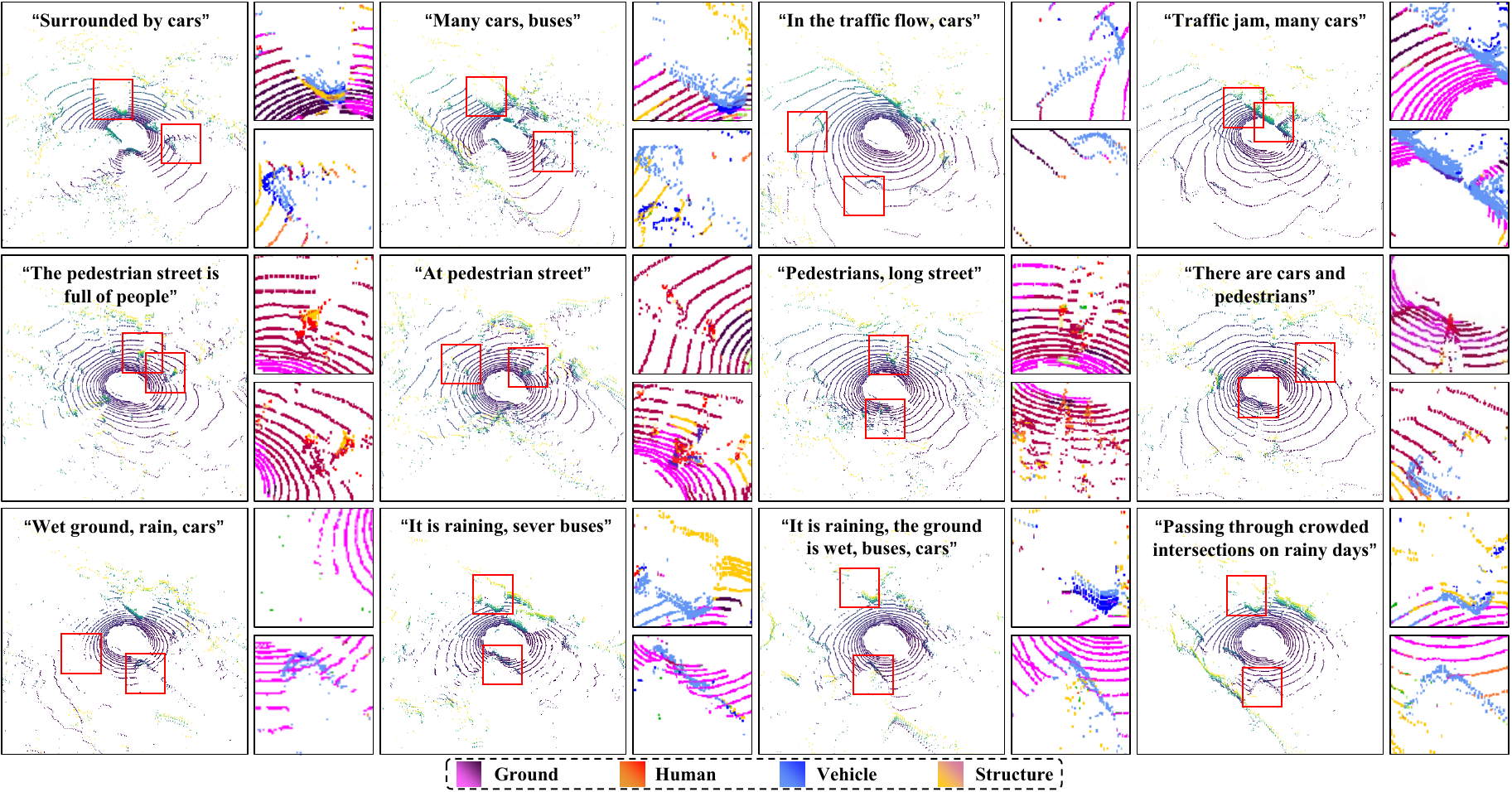}
\caption{Results of text-controlled LiDAR point clouds generation.}
\label{fig: text}
\end{figure}
Previous methods are all limited to unconditional generation and densification, only our approach enables text-controlled generation, which can further expand application scenarios.
Thanks to our designed CEI, our model can effectively integrate textual semantics into the generation process.
Figure~\ref{fig: text} illustrates the text-controlled 32-beam generations.
To verify the semantic accuracy of the generated objects, we employ RangeNet~\cite{milioto2019rangenet} for semantic segmentation, and the zoomed-in results are displayed in the right column.
Additional diverse text-controlled generation results are also showcased in the supplementary materials

\noindent\textbf{Larger Objects Generation.}
It can be observed from the first row that the generated outlines of vehicles are very clear, and the model is capable of generating various forms of vehicles, including small cars, buses, and trucks. The distribution of LiDAR beams also aligns with the characteristics of real data.

\noindent\textbf{Human Generation.}
Generating humans is particularly challenging, as the LiDAR points that successfully reflect off and are detected on a human body in an outdoor scene are quite sparse. It can be observed from the second row that our method is capable of generating high-quality representations of humans while maintaining the coherence of other parts of the point cloud.

\noindent\textbf{Rainy Day Generation.}
The third row shows the more challenging and meaningful results of generating LiDAR point clouds on rainy days. Due to the wet surfaces of the ground and objects from rain, the beams illuminating them experience total reflection, making it challenging for LiDAR to detect them. 
This leads to sparser or even zero returning beams from targets that are farther away from the LiDAR. 
%
%
Particularly noteworthy is the third example in the third row, which illustrates a typical phenomenon where beams that illuminate the vehicle can return, while those illuminating the ground around the vehicle cannot.

\subsection{Ablation Study and Analysis}
\begin{table}[!t]
\centering
\caption{Results of ablation study on key designs.}
\resizebox{0.95\textwidth}{!}{
\begin{tabular}{cccccccccc}
\bottomrule
\multirow{2}{*}{Versions} & \multicolumn{5}{c}{Key designs} & Point cloud & Image  & \multicolumn{2}{c}{BEV occupancy grid} \\ \cmidrule(r){2-6}\cmidrule(r){7-7}\cmidrule(r){8-8}\cmidrule{9-10}
& EA & REA & CEI & FM & D-layers & FPD\scriptsize$\downarrow$ & FRD\scriptsize$\downarrow$\qquad & MMD\scriptsize$\times10^{-4}\downarrow$  & JSD\scriptsize $\times10^{-2}\downarrow$  \\ \hline
Text2LiDAR$_A$ & - & - & - & - & 3 & 25.03 & 692.15 & 1.26 & 3.82 \\
Text2LiDAR$_B$ & $\checkmark$ & - & $\checkmark$ & $\checkmark$ & 4 & 8.02  & 305.27 & 0.81 & 2.90 \\
Text2LiDAR$_C$ & - & $\checkmark$ & $\checkmark$ & $\checkmark$ & 4 & 7.41  & 220.11 & 0.87 & 3.09 \\
Text2LiDAR$_D$ & - & - & $\checkmark$ & $\checkmark$ & 4 & 10.52 & 416.03 & 0.89 & 3.30 \\
Text2LiDAR$_E$ & $\checkmark$ & $\checkmark$ & - & $\checkmark$ & 4 & 6.99  & 187.82 & 0.83 & 2.79 \\ 
Text2LiDAR$_F$ & $\checkmark$ & $\checkmark$ & $\checkmark$ & - & 4 & 5.95  & 165.33 & 0.52 & 2.45 \\
Text2LiDAR$_G$ & $\checkmark$ & $\checkmark$ & - & - & 4 & 11.15 & 333.69 & 1.17 & 4.18 \\
Text2LiDAR$_H$ & $\checkmark$ & $\checkmark$ & $\checkmark$ & $\checkmark$ & 3 & 19.36 & 521.32 & 0.97 & 3.59 \\
\textbf{Text2LiDAR} & $\checkmark$ & $\checkmark$ & $\checkmark$ & $\checkmark$ & 4 & \textbf{4.81}  & \textbf{164.16} & \textbf{0.49} & \textbf{2.01} \\ 
        \bottomrule
\end{tabular}}
\label{tab: ablation}
\end{table}
We conducted experiments on KITTI-360, varying the effectiveness of the main designs in this paper, including EA, REA, CEI, FM, and the number of decoder layers (D-layers). From Table~\ref{tab: ablation}, we can see the experimental results.
%

\noindent\textbf{Benefits of (Reverse) Equirectangular Attention.}
Table~\ref{tab: ablation} indicates that when we do not use EA or REA and only use the original self-attention as a substitute, the model's performance deteriorates. The performance drop is more pronounced when REA is not used, as upsampling is closer to the model's endpoint, directly impacting high-resolution output. When both REA and EA are not used, the model's performance further declines, deteriorating by $5.71, 251.87, 0.4\times10^{-4}$, and $1.29\times10^{-2}$,  in the respective four metrics.
Meanwhile, we find that increasing the number of decoding layers is an effective way to enhance performance, further confirming the effectiveness of REA.

\noindent\textbf{Benefits of Control-signal Embedding Injector.}
In terms of feature fusion, we compare channel-wise concatenation with the designed CEI. We can see from Text2LiDAR$_E$ that, in addition to fusing text embedding, the designed injector is also more effective in incorporating timestep embedding, ensuring the stability of the reverse process.

\noindent\textbf{Benefits of Frequency Modulator.}
From Text2LiDAR$_F$, we can find that, 
without FM, all four metrics show varying degrees of decline: $0.78, 1.17, 0.03\times10^{-4}$, and $0.44\times10^{-2}$. Furthermore, we observe that FM has a relatively minor impact on FRD. We attribute this to the small proportion of high-frequency information, and adjusting it could introduce some image distortion, which is crucial for the performance of point clouds.

\begin{table}[!t]
\centering
\caption{Comparison with the parameters and generation speed of previous methods.}
\resizebox{0.9\textwidth}{!}{
\begin{tabular}{ccccc}
\bottomrule
Methods  & Network architecture & Control signal & Parameters\scriptsize(M) & Time\scriptsize(s) \\ \hline
LiDARGen~\cite{zyrianov2022learning} & RefineNet            & None & 29.7   & 88.54    \\
R2DM~\cite{nakashima2023lidar}     & Efficient U-Net      & None & 31.1   & 7.79    \\
Text2LiDAR(ours) & Transformer          & None & 45.8   & 12.57     \\ 
Text2LiDAR(ours) & Transformer          & Text & 46.1   & 4.57     \\
\bottomrule
\label{tab: parameter}
\end{tabular}}
\end{table}
\noindent\textbf{Parameter Analysis.}
Table~\ref{tab: parameter} shows the comparison of model parameters and generation speed between our method, R2DM~\cite{nakashima2023lidar}, and LiDARGen~\cite{zyrianov2022learning}. Our model has two versions, distinguished based on the control signal.
It can be observed that our method has a larger number of parameters, but this does not hinder the ability to perform fast generating. When generating 32-line LiDAR data with text control, our model achieves high speed, requiring less than five seconds. This indicates that our model is highly practical.

\section{Conclusion}
We have proposed Text2LiDAR, the first high-performance text-controllable LiDAR data generation framework.
With the designed EA, Text2LiDAR can specifically capture LiDAR features from equirectangular images while fully considering the physical meaning of equirectangular images.
Then, a CEI has been proposed to integrate control signals
, ensuring high-quality text-controlled generation.
We have further proposed an FM to overcome the issue of excessive smoothing in equirectangular images and assist the model training.
Our Text2LiDAR has shown great performance in uncontrolled and text-controlled generation, and densification, with promising applications across various domains.
At last, we have constructed the nuLiDARtext, providing the paired text-LiDAR data to facilitate advancements in this field.

\section*{Acknowledgements}
We are very grateful for the reviewers' critical and constructive comments. This work was supported in part by the NSFC under Grant No. 62276141, 62176124, 62276144, 62361166670.

\bibliographystyle{splncs04}
\bibliography{egbib}
\end{document}